\pdfoutput=1

\documentclass[11pt]{article}

\usepackage{acl}

\usepackage{times}
\usepackage{latexsym}

\usepackage[T1]{fontenc}

\usepackage[utf8]{inputenc}

\usepackage{microtype}

\usepackage[draft]{todonotes}   

\usepackage{xspace} 
\usepackage[capitalize,noabbrev]{cleveref} 
\newcommand{\dants}{DanT5\xspace}

\usepackage{multirow,booktabs, hhline} 
\usepackage[caption=false]{subfig}

\title{Training a T5 Using Lab-sized Resources}

\author{Manuel R. Ciosici \\
  USC Information Sciences Institute \\
  USA \\
  \texttt{manuelc@isi.edu} \\\And
  Leon Derczynski \\
  ITU Copenhagen \\
  Denmark \\
  \texttt{ld@itu.dk} \\}

\begin{document}
\maketitle

\begin{abstract}
Training large neural language models on large datasets is resource- and time-intensive. 
These requirements create a barrier to entry, where those with fewer resources cannot build competitive models.
This paper presents various techniques for making it possible to (a) train a large language model using resources that a modest research lab might have, and (b) train it in a reasonable amount of time. 
We provide concrete recommendations for practitioners, which we illustrate with a case study: a T5 model for Danish, the first for this language.
\end{abstract}

\section{Introduction}

Pre-training large language models (LLMs) is costly. There is an environmental cost~\cite{strubell-etal-2019-energy,patterson_carbon_2021}; money, in running costs~\cite{Sharir2020}; and potentially money in hardware, where the equipment to provide the needed computational power is often outside of a modest research lab's budget.

This paper presents two contributions.
The first is a set of recommendations to practitioners for training a modern LLM on modest hardware, limited data, and a reasonable compute time.
The second is a demonstration of the recommendations resulting in \dants, a T5 model for Danish, the first for this language.\footnote{DanT5-small and DanT5-large are available at  \url{https://huggingface.co/strombergnlp/dant5-small} and \url{https://huggingface.co/strombergnlp/dant5-large}.}

\section{Computational challenges}
    
\textbf{GPU memory} is the most significant hurdle for training LLMs. State-of-the-art optimizers like \texttt{Adam}~\cite{kingma_adam_2015,loshchilov_decoupled_2018} converge much faster than traditional \texttt{Stochastic Gradient Descent} due to tracking the first and second order momentum, but the performance comes at a cost. However, to track the momentum, \texttt{Adam} must keep in memory two additional values for each parameter in the model, thus adding a 2X memory overhead.

Recent research has investigated \textbf{smaller-than-default number representations}, taking a floating point value (typically 32 bits) down to 16, 8, or fewer bits~\cite{NIPS2017_b1a59b31,wang2018training,kalamkar_study_2019}. While these advances have the potential to ease and so democratize large model training, they require hardware and software support, which is still ``just over the horizon".

Using \texttt{float16} representations typically reduces a model's GPU memory consumption by half~\cite{micikevicius_mixed_2018} and require a relatively modern GPU and software stack.\footnote{NVIDIA Pascal architecture GPU or newer, CUDA 8+. Modern PyTorch or TensorFlow.} However, these reduced-range representations can overflow if training a model that expects a higher dynamic range. This can happen, for example, when fine-tuning a model initially trained using TPUs, which default to \texttt{bfloat16} for float representation. 
 
The \texttt{bfloat16} representation should reduce compute needs~\cite{kalamkar_study_2019}, though using it requires hardware\footnote{An NVIDIA Ampere GPU or newer.} and software support, and implementations across common software frameworks are still ``not quite finished just yet". Using even coarser representations, such as \texttt{bitsandbytes}~\cite{dettmers_8-bit_2021} does, should also help, but these are just beginning to be implemented in popular NLP frameworks.\footnote{For example, the Hugging Face \texttt{transformers} library only added support for 8-bit optimizers starting with version $4.19$, released on May 12, 2022.}

One approach to training LLMs with modern GPUs is model parallelism, where a language model is split over multiple GPUs, with each GPU storing one or more layers. However, model parallelism brings the challenge of \textbf{data communication}. GPU-to-GPU communication typically goes through the slow PCI Express bus and sometimes the even slower CPU-to-CPU NUMA interconnect. Communication effects can be ameliorated via hardware and software. In hardware, one can use NVLink bridges to support high-speed GPU-to-GPU communication. NVLink bridges are a cost-effective solution for linking GPU pairs in a machine. NVLink bridges have a bandwidth of 50-100GB/s (depending on the GPU generation) compared to 31.5 GB/s for PCIe 4.0. On the software side, \texttt{DeepSpeed}~\cite{Ren2021,9355301,rajbhandari_zero-infinity_2021} allows complex GPU offloading --  supporting, for example, moving the expensive \texttt{Adam} optimizer from GPU memory and into (much larger) CPU memory. \texttt{DeepSpeed} CPU optimizers are implemented in efficient C++ code that uses the CPU's native SIMD support. While offloading the optimizer to system memory increases the system memory requirements, RAM capacity is usually extendable at a reasonable cost. GPU memory is not extendable.

Given the challenges discussed above, we recommend the following when choosing compute hardware for training Large Language Models:
\begin{itemize}
    \item Prioritize GPU memory over computation speed. LLM training is usually limited by: the effects of low memory; communication bandwidth between GPUs; and gradient accumulation over small batches.
    \item Install NVLink bridges to speed up GPU-to-GPU communication. Modern ML frameworks automatically use NVLink for GPU-to-GPU data transfer.
    \item Prioritize GPUs with support for \texttt{float16} and \texttt{bfloat16} in order to take advantage of optimizations whose widespread support is very close/nascent.
    \item Plan for system memory of 512GB or more. If the budget is tight, buy larger capacity RAM modules and leave empty RAM slots for future upgrades. 
    \item Leaving PCIe slots for future GPU acquisitions can extend the life of a lab's computing machines. For example, buying computers with support for eight GPUs even though budget only permits four, allows for staggered GPU acquisition.
\end{itemize}

\section{Working with sub-gargantuan corpora}
\label{sec:working_with_small_corpus}

There is a large disparity in the size of corpora available for English and for other languages.
While English has, for example, ``The Pile''~\cite{gaoPile800GBDataset}, an automatically-gathered corpus of over 800GiB, most other languages have to get by with datasets orders of magnitude smaller~\cite{dunn-adams:2020:LREC}.

Apart from size, corpus quality matters.
Large corpora, such as the Common Crawl, include significant amounts of undesirable/abusive/illegal material~\cite{luccioni-viviano-2021-whats,birhane2021multimodal}, which leads to harmful model behavior~\cite{10.1145/3442188.3445922b}; web-spider-derived corpora also tend to duplicate data, requiring de-duplication, a deceptively complex and often challenging task~\cite{lee_deduplicating_2021}.
While these issues may have partly reduced impact with massive corpora, the effect of each unwanted piece of content is likely to be larger the smaller the corpus.
When dealing with corpora of a few gigabytes or less, quality becomes much more important, and the advantages of automatic data collection can wane.

As a case study, we look at building a model for Danish using the T5~\cite{Raffel2019a} architecture, \dants.
Danish is a modestly-resourced language~\cite{kirkedal-etal-2019-lacunae}.
The original T5 used 34B sentence piece tokens from C4.
The Norwegian T5,\footnote{\url{https://huggingface.co/NbAiLab/nb-t5-base-v2}} for a language similar to Danish, used the \emph{Norwegian Colossal Corpus}~(NCC, \citet{kummervold-etal-2021-operationalizing}, 7B words\footnote{\url{https://github.com/NBAiLab/notram}}).
Unfortunately, there is no high-quality Danish language text corpus the size of C4 or even the NCC.

The largest freely available high-quality Danish corpus is the \emph{Danish Gigaword Corpus} (DAGW, \citet{stromberg-derczynski-etal-2021-danish}), a curated 1B word corpus. 
Like the NCC, DAGW assembles text spanning several domains, dialects, time periods, and modalities.

A single training pass over DAGW is too little training data for a modern LLM, even when warm-starting the training~(see \Cref{sec:warm_starting}). 
To compensate for this, when training \dants, we perform 10 epochs over the DAGW with \emph{dynamic masking}.
Before training, we split the corpus into sequences of 512 word pieces, but do not mask any token.
During training, when each batch is assembled, we randomly mask 15\% of each sequence's tokens.
The random, just-in-time masking results in different masks for each epoch. 
Exposing \dants to constantly changing masks attempts to compensate for DAGW's size by generating multiple masked configurations for any given sequence in the source corpus. We mask 15\% of the tokens in each sequence, following T5's original training procedure.

\section{Warm starting from a different language}
\label{sec:warm_starting}
    

Warm-starting is a shortcut to training a large language model~\cite{rothe_leveraging_2020}. Warm-starting does not train a model from scratch but continues training a pre-existing LLM. Therefore, warm-starting reduces computation costs and environmental impact by taking advantage of some of the computation that went into the source LLM.  We used the original T5 checkpoints~\cite{Raffel2019a} as the basis for \dants. Warm-starting \dants from an English T5 also supports transfer learning as the two languages are somewhat related.\footnote{English is a West Germanic language, while Danish is North Germanic}.

Language models trained from pre-existing models are constrained to using the source model's tokenizer and, therefore, its vocabulary of tokens. 
The English T5's vocabulary contains sub-word tokens representing mostly English, with a minority of tokens specialized for Romanian, German, and French.\footnote{These tokens are present in the original T5 to allow the model to perform machine translation for some language pairs.} Reusing the sub-word token vocabulary for Danish works but introduces excessive word splitting and leaves many tokens rarely used. Splitting words excessively results in long input sequences, reducing the amount of content that can fit T5's maximum input length of 512. Rarely-used tokens occupy space in the embedding layer, displacing tokens relevant to the target language.

Several approaches have been proposed to modify a pre-trained language model's vocabulary. But, most methods cannot convert a mostly-monolingual vocabulary into another mostly-monolingual vocabulary, as needed for \dants. For multi-language machine translation models, \citet{garcia_towards_2021-1}~extend a highly-multilingual vocabulary with tokens for a new language, but assume that the vocabulary grows by only a tiny fraction. \citet{chronopoulou_reusing_2020}~first train a language model using a tokenizer trained only on the source language. Then they train a tokenizer on the concatenation of the source and target language corpora. The target language model then reuses word embeddings for the tokens appearing in both tokenizers. This approach cannot support adapting a pretrained language model for warm-start training in a new language. More recently,  \citet{10.1162/tacl_a_00461} proposed doing away with tokenizers altogether and training T5 models directly on byte encodings. This method promises token-free models, but the requirement for a deeper encoder and the 30\% larger training cost push this approach even further away from what can be achieved with the resources of a modest research lab.

Below, we present a new but simple way to adapt the vocabulary and parameters of English T5 to create a good (warm-)starting point for training \dants. Our approach can generate warm-starting models for any language and transformer architecture (encoder-only, decoder-only, or encoder-decoder, like T5).

\textit{Warm-start tokenization translation.} We first follow \citet{Raffel2019a} and train a mostly Danish sentence-piece tokenizer on a mixture of 90\% Danish and 10\% English. We include English for the same reason as \citet{Raffel2019a}: so that we can later fine-tune the model to support machine translation, in this case Danish-English. \dants's tokenizer has exactly the same size as T5's to make the warm starting process easier, but one could extend or reduce the vocabulary as needed. 
For warm-starting, we adapt the original T5 to the Danish tokenizer using a simple approach. First, we attempt to translate into English every token in the Danish tokenizer using Google Translate. We record the successful translations, and where the translation fails, we set the translation to equal the Danish word. We then create the starter \dants model by cloning T5 and adjusting the input embeddings.\footnote{T5's input and output embeddings are identical, so we effectively adjust both the input and output embeddings.} We set each embedding in \dants to the mean of the English embeddings of the English translation. \Cref{tbl:example_token_mapping} shows examples of Danish words translating to single or multiple English words and the contingency in case of translation failure (last row).

\begin{table}[t]
    \centering
    \begin{tabular}{ l|l|l }
        \textbf{Danish} & \textbf{English} & \texttt{en} \textbf{tokens} \\
        \midrule
        doktor & doctor & doctor\\
        dokumentet & the document & the; document\\
        værsgo & here you go & here; you; go\\
        Yndling & Favorite & Favor; it; e \\
        Aarhus & ERROR! & A; ar; hus
    \end{tabular}
   \caption{Example Danish to English token mappings. The last row shows our approach for cases where translation fails.}
\label{tbl:example_token_mapping}
\end{table}

\section{\dants Training}
\label{sec:training_details}

We follow T5's original training procedure~\cite{Raffel2019a} with a few exceptions. When we trained \dants, there was no \texttt{bfloat16} support in DeepSpeed or \texttt{transformers}, so we trained our models with \texttt{float32} representations.

\emph{Optimizer.} We use DeepSpeed's \texttt{AdamW} implementation instead of \texttt{Adafactor}~\cite{Raffel2019a,shazeer_adafactor_2018}. We use a learning rate of $4e-3$, with warmup over the first $5\ 000$ steps and a linear learning rate decay for the remaining. We chose \texttt{AdamW} over \texttt{Adafactor} due to the availability of a CPU-offloadable implementation of \texttt{AdamW} through DeepSpeed, an essential requirement when constrained by GPU memory. Even so, to achieve a batch size of $128$, we accumulate gradients over $8$ forward passes.

\emph{Batch Packing.} Unlike T5's original training, we do not pack batches with multiple sequences to obtain batches of roughly the same number of tokens. Instead, we dynamically pad each batch so that all inputs in a batch are padded to the maximal length within the batch. Therefore, length varies across batches but minimizes padding within each batch. As opposed to static padding, this technique speeds up computation on GPUs as it reduces the time wasted computing activations for pad tokens. While batch packing can be adapted to GPU computation, it is best suited for computations of TPUs which do not support dynamic padding.

\emph{Dynamic Masking.} The original T5's training consumed 1 trillion tokens from the \emph{Colossal Clean Crawled Corpus (C4)}, a procesed subset of \emph{Common Crawl}. Since no equivalently large corpus exists for Danish, we used the \emph{Danish Gigaword Corpus (DAGW, \citet{stromberg-derczynski-etal-2021-danish})} together with \emph{dynamic masking} (see \Cref{sec:working_with_small_corpus}).

\emph{Hardware.} We train \dants on a single machine with two AMD Epyc 7252 8-Core CPUs, 128 GB of RAM, and four NVIDIA A100 GPUs with 40 GB of memory each. We did not have NVLink bridges for this project.

\section{Case Study Results}
\label{sec:results}

We trained two models: \texttt{dant5-small}, a 60M parameter model based on  \texttt{T5-small}, and \texttt{dant5-large}, a 770M parameter model based on \texttt{T5-large}. Both models trained for 10 epochs, as described in \Cref{sec:working_with_small_corpus}. Training \texttt{dant5-small} took 91 hours, while \texttt{dant5-large} trained in 508 hours.

\begin{figure}[htp]
    \centering
    \subfloat[Training loss]{%
        \includegraphics[clip,width=0.85\columnwidth]{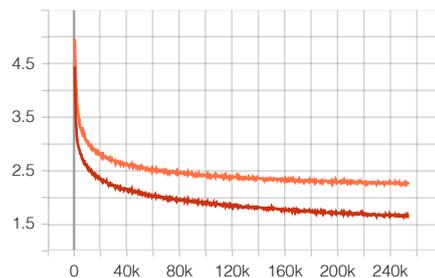}%
        \label{fig:training_loss}
    }
    
    \subfloat[Evaluation loss]{%
        \includegraphics[clip,width=0.85\columnwidth]{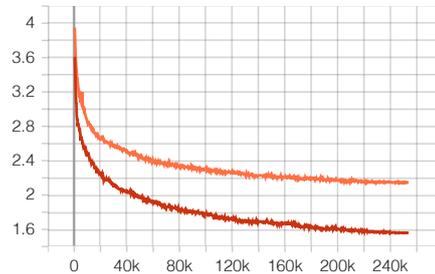}%
         \label{fig:evaluation_loss}
    }
    
    \caption{Training and Evaluation loss for \texttt{dant5-small} (orange, 60M parameters) and \texttt{dant5-large} (red, 770M parameters).}
    
\end{figure}

\Cref{fig:training_loss} indicates that \texttt{dant5-large} fit the training data better. The training loss decreases up to the end, suggesting that the models are still learning. The evaluation loss curve (shown in \Cref{fig:evaluation_loss}) closely tracks the training loss, indicating that the models are not overfitting the data.

\section{Conclusion}
\label{sec:conclusion}

This paper presents a set of concrete recommendations to enable training LLMs using modest research lab resources, in a reasonable amount of time. We provide parameter values, software and hardware configuration strategies, and techniques for enhancing the use of available data. These recommendations are then demonstrated in the creation and release of ``\dants", the first T5 model for Danish.

\section*{Acknowledgements}
This work was conducted with support from  the Independent Danish Research under the project VerifAI, 9131-00131B, and the Novo Nordisk Foundation project ClinRead, NNF19OC0059138.

\bibliography{custom}



\end{document}